\documentclass[conference]{ieeeconf}

\IEEEoverridecommandlockouts                            

\usepackage{graphics} 
\usepackage{epsfig} 
\usepackage{mathptmx} 
\usepackage{times} 
\usepackage{amsmath} 
\usepackage{amssymb}
\usepackage{bbold}
\usepackage{multirow}
\usepackage{subcaption}
\usepackage{cite}

\title{\LARGE \bf
MIFair: A Mutual-Information Framework for Intersectionality and Multiclass Fairness
}

\author{Jeanne Monnier$^{1, 2}$, Thomas George$^{1}$, Christèle Tarnec$^{1}$, Frédéric Guyard$^{1}$, and Marios Kountouris$^{2}$
\thanks{$^{1}$Orange Research, Châtillon, France}%
\thanks{$^{2}$EURECOM, Sophia Antipolis, France}%
}

\begin{document}

\maketitle

\thispagestyle{empty}
\pagestyle{empty}

\begin{abstract}
Fairness in machine learning remains challenging due to its ethical complexity, the lack of a universal definition, and the need for context-specific bias metrics. Existing methods also remain limited in handling intersectionality, multiclass settings, and broader flexibility and generality. To address these limitations, we introduce \emph{MIFair}, a unified framework for bias assessment and mitigation based on mutual information. MIFair provides a flexible metric template and an in-processing mitigation method inspired by the Prejudice Remover, defining group fairness as statistical independence between prediction-derived variables and sensitive attributes, while establishing equivalences with widely used notions such as independence and separation.
MIFair naturally supports intersectionality, complex subgroup structures, and multiclass classification and employs regularization-based training to reduce bias according to the selected metric. Its key advantage is its versatility: it consolidates diverse fairness requirements into a single coherent framework, enabling consistent benchmarking and facilitating practical adoption.
Experiments on real-world tabular and image datasets show that MIFair effectively reduces bias, including in previously unaddressed multi-attribute scenarios, while maintaining strong predictive performance across all evaluated settings.
\end{abstract}

\section{Introduction}\label{sec:intro}
Fairness in AI has received increasing attention because biased models can cause harm in high-stakes applications. However, fairness in machine learning remains a broad and contested concept, which has led to multiple, often incompatible, fairness notions \cite{makhlouf2021machine, mitchell2018prediction}. Moreover, AI systems can reproduce or amplify biases present in the training data, and such biases may arise throughout the development pipeline, from data collection to model deployment and user feedback~\cite{mehrabi2021survey}. As AI systems become more influential, mitigating these biases is increasingly important. Yet most fairness metrics remain tied to specific formulations, limiting their generality and their ability to reflect the diversity of real-world settings.

Fairness in AI/ML is commonly studied through individual, group, and causal perspectives; here, we focus on group fairness. Group fairness includes multiple criteria \cite{makhlouf2021machine}, each capturing a different notion of fairness across demographic groups \cite{ruf2021towards, barocas2023fairness}. Beyond standard notions such as Statistical Parity, we also consider Conditional Fairness \cite{xu2020algorithmic}, which defines fairness as the conditional independence of model outcomes and sensitive attributes. Building on this view, MIFair unifies a broader set of fairness criteria within a mutual-information framework while supporting intersectional and multiclass settings, enabling more comprehensive and realistic fairness evaluation.

Information-theoretic approaches to measuring and enforcing fairness have gained prominence, building on foundational work such as \cite{kamishima2012fairness}, which we extend in several directions. Many conditional-fairness methods likewise rely on information-theoretic tools, including Shannon entropy and conditional mutual information \cite{ghassami2018fairness, cho2020fair}, as well as Jensen–Shannon divergence \cite{hwa2024enforcing}.

Our work builds on in-processing mitigation methods \cite{zhang2018mitigating, madras2018predict, iosifidis2019adafair, krasanakis2018adaptive, jiang2020identifying, jiang2020wasserstein, do2022fair, ghassami2018fairness, hwa2024enforcing} by extending regularization-based approaches to address biases arising from both the data and the training process. The closest prior works are \cite{kamishima2012fairness} and \cite{cho2020fair}. Prejudice Remover \cite{kamishima2012fairness} uses mutual information to enforce Statistical Parity, but it neither goes beyond this criterion nor explicitly evaluates intersectionality. Cho et al. \cite{cho2020fair} extend this line to Equalized Odds, yet still focus only on these two criteria and do not experimentally address intersectionality. Both methods also differ from ours in how mutual information is approximated and integrated into training, and neither considers multiclass classification.

In this paper, we introduce MIFair (pronounced “Am I Fair?”), an information-theoretic framework for unified fairness evaluation and mitigation in AI/ML. MIFair reformulates group fairness through mutual information, extending the Prejudice Remover \cite{kamishima2012fairness} into a flexible metric template that captures multiple fairness notions, together with a corresponding regularization-based in-processing method. A key contribution of MIFair is its explicit support for two important yet underexplored challenges: \emph{intersectionality} and \emph{multiclass classification}. While most prior work focuses on single-attribute or binary settings \cite{yang2020fairness}, and existing tools rarely support intersectionality in practice despite its importance for capturing structural discrimination \cite{buolamwini2018gender}, MIFair enables fairness assessment and mitigation across complex subgroup structures and multiclass prediction tasks \cite{celis2019improved, kamiran2012data}. Experiments on Adult and CelebA show that MIFair effectively reduces bias in intersectional and multiclass settings while providing a unified information-theoretic framework for comparing fairness notions. By reducing the fragmentation of existing fairness methods, MIFair offers a more accessible, adaptable, and principled approach to fairness in machine learning.

\section{Problem setting} \label{sec:pb}

\textbf{Data and sensitive attributes.}
Let $D=\{(\mathbf{x}_d,\mathbf{a}_d,y_d)\}_{d=1}^{|D|}$ be a dataset drawn from an underlying distribution over $(\mathbf{X},\mathbf{A},Y)$, where each sample represents an individual. Here, $\mathbf{x}_d=(x_{d,1},\ldots,x_{d,m}) \in \mathcal{X}$ denotes the vector of non-sensitive attributes, $\mathbf{a}_d=(a_{d,1},\ldots,a_{d,n-m}) \in \mathcal{A}$ denotes the vector of sensitive attributes, and $y_d \in \mathcal{Y}$ denotes the prediction target. The sensitive attributes are collected in the random vector $\mathbf{A}=(A_1,\ldots,A_{n-m})$, with sensitive-attribute space $\mathcal{A}=\prod_{i=1}^{n-m}\mathcal{A}_i$; these are the features with respect to which discrimination should be avoided, and their selection is context- and application-dependent. The remaining features are collected in the random vector $\mathbf{X}=(X_1,\ldots,X_m)$, with feature space $\mathcal{X}=\prod_{i=1}^{m}\mathcal{X}_i$, and are referred to as non-sensitive attributes. The number of sensitive attributes is $n-m$. Although these variables may in principle be discrete or continuous, this paper focuses on discrete classification settings; accordingly, the mutual-information expressions are written in discrete form.

\textbf{Demographic groups.}
Given the sensitive attributes, we define \textit{demographic groups} as sets of individuals sharing the same joint sensitive-attribute value $\mathbf{a} \in \mathcal{A}$. Let $G$ denote the set of such groups, and let $|G|$ denote its cardinality. We distinguish two settings:
\begin{enumerate}
    \item $|G| = 2$, corresponding to the binary case, which is the setting most commonly studied in the literature;
    \item $|G| > 2$, corresponding to the non-binary case, which is more realistic in practice but far less explored.
\end{enumerate}
\textbf{Intersectionality.}
Intersectionality acknowledges that discrimination often arises from the combination of multiple sensitive attributes rather than from any single one considered in isolation. When multiple sensitive attributes are considered jointly, the setting typically becomes non-binary and may yield $|G| > 2$ demographic subgroups. Real-world applications commonly involve such intersectional structures, requiring fairness metrics capable of handling vectors of sensitive attributes.

\textbf{Fairness metrics.}
Fairness is an ethical principle that cannot be captured by a single universal rule, which has led to the development of multiple \textit{fairness notions}. Mathematical formalizations in AI/ML therefore yield diverse, and often incompatible, criteria \cite{mitchell2018prediction}. Selecting an appropriate notion is therefore crucial to ensure alignment with the data and application context and to avoid unintended or amplified biases.
An extensive overview of fairness notions is provided in \cite{makhlouf2021machine}, with a subset shown in Table~\ref{tab:metrics_sota}. These criteria are typically defined in a simplified setting in which both the sensitive attribute $\mathbf{A}$ and the model output $\hat{Y}$ are binary, although some notions extend to multiclass classification or regression. Most definitions rely on the joint distributions of $\mathbf{A}$, $Y$, and $\hat{Y}$. Following this line of work, we focus on three well-known fairness notions and two relaxations introduced below (see Section~\ref{subsec:templatesotametric}), covering a broad portion of the existing literature.
\begin{table*}[h]
\centering
\resizebox{\textwidth}{!}{%
\begin{tabular}{|c|c|c|c|}

\hline Fairness Notion & Formulation & Ref. & Classification \\
\hline \hline Statistical parity & $P(\hat{Y} \mid A=0)=P(\hat{Y} \mid A=1)$ & \cite{dwork2012fairness} & Independence \\
\hline
Equalized odds & $P(\hat{Y}=1 \mid Y=y, A=0)=P(\hat{Y}=1 \mid Y=y, A=1) \quad \forall y \in\{0,1\}$ & \multirow{2}{*}{\cite{hardt2016equality}} & \multirow{5}{*}{Separation} \\
\cline{0-1}  Equal opportunity & $P(\hat{Y}=1 \mid Y=1, A=0)=P(\hat{Y}=1 \mid Y=1, A=1)$ & & \\
\cline{0-2} Predictive equality & $P(\hat{Y}=1 \mid Y=0, A=0)=P(\hat{Y}=1 \mid Y=0, A=1)$ & \cite{corbett2017algorithmic} & \\
\hline Overall accuracy equality & $P(\hat{Y}=Y \mid A=0)=P(\hat{Y}=Y \mid A=1)$ & \cite{berk2021fairness} & Other metrics \\
\hline
\end{tabular}}
\caption{Classification of state-of-the-art fairness notions \cite{makhlouf2021machine}}
\label{tab:metrics_sota}
\end{table*}

\textbf{Model training and optimization problem}.
Our goal is to learn a fair model by modifying the training objective to reduce bias while preserving the task’s essential structure. Let $\mathbf{w}$ denote the model parameters. Standard learning seeks to approximate the ground-truth distribution $P(Y | \mathbf{X}, \mathbf{A})$ through Empirical Risk Minimization (ERM):
\begin{equation}\label{pbopti}
\min_{w \in \mathbb{R}^{h}} L(D,w) = \min_{w \in \mathbb{R}^{h}} \sum_{d=1}^{|D|} \ell(\mathbf{x}_d,\mathbf{a}_d,y_d;w),
\end{equation}
where $L$ is the empirical risk and $\ell(\cdot)$ denotes the loss incurred on instance $d$. 
The ERM fits the patterns present in the data; therefore, if the data are biased, the trained model will tend to reproduce these biases, making the standard ERM insufficient for learning fair models.

\section{MIFair: A Versatile Framework for Fairness Metrics} \label{sec:template}
We now introduce the bias-assessment component of MIFair: a versatile metric template that unifies fairness evaluation and accommodates the diversity of existing methods. This template broadens the applicability of group fairness metrics by providing a standardized and extensible formulation\footnote{Throughout this work, we use raw mutual information $I(A;B)$ as the fairness criterion and optimization quantity. When cross-notion comparison on a common numeric range is desired, a normalized variant such as $I(A;B)/H(B)$, or its conditional analogue $I(A;B\mid C=c)/H(B\mid C=c)$, may also be required whenever the denominator is nonzero. Accordingly, MIFair should be interpreted primarily as a unified information-theoretic formulation; raw MI values across different fairness notions are not, in general, directly comparable on a strictly common scale.} for the major fairness notions. We present the definitions used to measure bias under these notions and show their equivalence to the formulations listed in Table~\ref{tab:metrics_sota}.

A close examination of group fairness metrics reveals a common underlying structure: each metric specifies a particular variable, defined as a function of the model’s predictions, that is directly tied to what is considered a non-prejudice (fair) treatment under the corresponding fairness notion. This variable can be interpreted as quantifying the \emph{benefit} (or, conversely, the detriment) of belonging to a particular demographic subgroup. It therefore provides a basis for comparing the model’s effects across subgroups. In the ideal case, a fully non-discriminatory model is obtained when the benefit variable is statistically independent of the sensitive attributes.

Measuring model bias can thus be framed as quantifying the mutual dependence between the sensitive attributes and the benefit variable derived from the model’s predictions. Building on this insight, as well as on the Prejudice Remover method of \cite{kamishima2012fairness}, we introduce an information-theoretic fairness metric template grounded in Mutual Information. This template accommodates a broad range of group fairness notions and naturally supports both intersectional analyses and multiclass classification settings, thanks to the flexibility of mutual information in handling random variables of different dimensions.

\subsection{A Metric Based on Mutual Information} \label{subsec:mi}
Building on the preceding observation, we quantify the \emph{mutual dependence} between the sensitive-attribute vector $\mathbf{A}$ and the benefit $B = f_b(\hat{Y}, Y)$, whose definition depends on the fairness notion under consideration (Section~\ref{subsec:templatesotametric}).

Using mutual information allows us to overcome two key limitations of existing fairness metrics. 
First, \textit{intersectionality}: most common metrics compare probabilities across only two groups, restricting their applicability to binary protected attributes. Although extensions exist \cite{kearns2018preventing}, they require numerous pairwise comparisons, which scale poorly as $|G|$ increases. Second, \textit{multiclass classification}: real-world tasks frequently involve more than two outcome classes, requiring fairness metrics that go beyond binary positive/negative predictions. 
Our formulation naturally addresses both challenges.

We leverage tools from information theory, namely Mutual Information (MI), as the foundation of our approach. MI is non-negative and equals zero if and only if the variables are statistically independent; lower values therefore indicate weaker statistical dependence.

We define our metric template as
\begin{equation}\label{MIeq}
\iota_{\mathrm{fairness}} = I(\mathbf{A};B) = \sum_{\mathbf{a}\in\mathcal{A}}\sum_{b\in\mathcal{B}}P_{\mathbf{A},B}(\mathbf{a},b)
\log\!\left(\frac{P_{\mathbf{A},B}(\mathbf{a},b)}{P_{\mathbf{A}}(\mathbf{a})\,P_B(b)}\right),
\end{equation}
where $P_{\mathbf{A},B}$ denotes the joint distribution of $\mathbf{A}$ and $B$, and $P_{\mathbf{A}}$ and $P_B$ denote their corresponding marginal distributions.

$I(\mathbf{A};B)$ quantifies the amount of information about the sensitive attributes that is encoded in $B$. A value of $0$ indicates complete fairness under the chosen notion, whereas larger values reflect stronger dependence and, therefore, greater bias.
In practice, the true distributions are unknown and are replaced by their empirical estimates, $\widetilde{P}_{\mathbf{A},B}$, $\widetilde{P}_{\mathbf{A}}$, and $\widetilde{P}_{B}$, which enables the computation of $\iota_{\mathrm{fairness}}$ in real-world scenarios (Section~\ref{sec:frame}).

\subsection{Unifying Existing Fairness Notions by Explicitly Defining the Corresponding Benefit} \label{subsec:templatesotametric}
We examine the five notions of fairness listed in Table~\ref{tab:metrics_sota} and, for each of them, provide an explicit instantiation of our metric template $\iota_{\text{fairness}}$ by specifying the corresponding benefit $B$. We also show that fairness under our metric implies fairness under the corresponding classical fairness metric.

Among these notions, Overall Accuracy naturally extends to multiclass classification. Independence- and Separation-based notions, originally defined for positive predictions in binary settings, can also be generalized either by focusing on a particular class or by requiring the distribution of the considered event ($\hat{Y}$ or $\hat{Y}|Y$) to be equal across all classes. Under this extension, Statistical Parity, Equalized Odds, and its relaxations, Equal Opportunity and Predictive Equality, also apply to multiclass scenarios. Our equivalence and implication results remain valid in this setting, illustrating the flexibility of the framework and its potential to define new categories of fair models.

\textbf{Statistical Parity.}
For Statistical Parity, the benefit $B$ is defined as the model prediction $\hat{Y}$, i.e., $B:=\hat{Y}$. The resulting fairness metric is
\begin{equation}
\iota_{SP} = I(\mathbf{A};\hat{Y}).
\end{equation}

\textbf{Equal Opportunity.}
Let $P_{Y=1} = P(\cdot \mid Y=1)$ denote the conditional distribution. Under Equal Opportunity, we measure the dependence between $\mathbf{A}$ and $\hat{Y}$ conditioned on $Y=1$, that is, we define $B := \hat{Y}\mid Y=1$:
\begin{align}
\iota_{EO} &= I_{Y=1}(\mathbf{A};\hat{Y}) = I(\mathbf{A};\hat{Y}\mid Y=1) \\
&= \sum_{\mathbf{a}\in\mathcal{A}}\sum_{\hat{y}\in\mathcal{Y}}
P_{Y=1}(\mathbf{a},\hat{y})\log\!\left(\frac{P_{Y=1}(\mathbf{a},\hat{y})}{P_{Y=1}(\mathbf{a})\,P_{Y=1}(\hat{y})}
\right).
\end{align}

\textbf{Predictive Equality.}
Let $P_{Y=0} = P(\cdot \mid Y = 0)$ denote the conditional distribution. Under Predictive Equality, we measure the dependence between $\mathbf{A}$ and $\hat{Y}$ conditioned on $Y = 0$, that is, we define the benefit variable as $B := \hat{Y}\mid Y=0$:

\begin{equation}
\iota_{PE} = I_{Y=0}(\mathbf{A};\hat{Y}) = I(\mathbf{A};\hat{Y}\mid Y=0).
\end{equation}

\textbf{Equalized Odds.}
Let $P_{Y=y} = P(\cdot \mid Y = y)$ denote the conditional distribution for $y \in \{0,1\}$. For Equalized Odds, we combine the conditional dependencies associated with both outcomes:

\begin{equation}
\iota_{EOdds} = \lambda_0 I_{Y=0}(\mathbf{A};\hat{Y})
+ \lambda_1 I_{Y=1}(\mathbf{A};\hat{Y}), \qquad \lambda_0,\lambda_1 > 0.
\end{equation}
where $\lambda_0, \lambda_1 > 0$ are weighting coefficients.

\textbf{Overall Accuracy Equality.}
For Overall Accuracy Equality, we define $B := \mathbb{1}\{\hat{Y} = Y\}$ and the corresponding metric becomes

\begin{equation}
\iota_{OAE} = I\!\left(\mathbf{A};\mathbb{1}\{\hat{Y}=Y\}\right).
\end{equation}

\section{MIFair: A Comprehensive Framework for Bias Mitigation} \label{sec:frame}
We introduce a bias mitigation framework grounded in our fairness metric $\iota_{\text{fairness}}$. MIFair operates as an in-processing method, addressing bias directly during model training. Among in-processing techniques, an effective way to improve fairness is to regularize the ERM objective in \eqref{pbopti}. Inspired by the Prejudice Remover \cite{kamishima2012fairness}, which our method generalizes, we incorporate our flexible metric $\iota_{\text{fairness}}$ as a tailored regularization term added to the loss function.
Standard ERM optimizes only for accuracy and therefore reproduces the biases embedded in the training data. Introducing a fairness regularizer shifts the objective toward approximating a modified distribution that remains close to the original distribution while exhibiting reduced dependence according to the selected fairness notion. This shift generally entails some accuracy loss on the training data, which is acceptable, or even desirable, when the data itself is biased. Nonetheless, maintaining adequate predictive performance remains essential. The trade-off between fidelity to the data and fairness is governed by the regularization hyperparameter $\eta$.
As discussed in Section \ref{subsec:mi}, minimizing mutual information reduces the dependence between two variables. Since our metric provides an empirical approximation of this quantity for the variables relevant to a chosen fairness notion, it serves naturally as a regularizer in the training objective. The initial optimization problem \eqref{pbopti} thus becomes

\begin{equation}\label{pb: final}
\min_{w\in\mathbb{R}^{h}} L(D,w) + \eta\,\iota_{\mathrm{fairness}}(D,w).
\end{equation}

The proposed framework consists of the following steps:
\begin{enumerate}
    \item \textbf{Identify sensitive attributes}: determine which features in the dataset should be treated as sensitive and define the associated demographic subgroups.
    \item \textbf{Select a fairness definition}: choose the appropriate notion for the task and define $\iota_{\text{fairness}}$ accordingly.
    \item \textbf{Set up the optimization problem}: incorporate the chosen $\iota_{\text{fairness}}$ into the training objective.
    \item \textbf{Train the model}: optimize the regularized objective.
\end{enumerate}

In practice, the value of $\iota_{\text{fairness}}$ required in \eqref{pb: final} is computed by estimating mutual information from minibatch samples. The joint and marginal distributions $P_{\mathbf{A},B}$, $P_{\mathbf{A}}$, and $P_B$ are approximated by their empirical counterparts, denoted by $\tilde{P}$, which are derived from the training data and the model output probabilities $p_{\mathbf{w}}$. Let $\mathcal{M}$ denote the current minibatch, with cardinality $|\mathcal{M}|$. The fairness regularizer is estimated from minibatch samples as

\begin{equation}
\hat{\iota}_{\mathrm{fairness}} = \sum_{\mathbf{a}\in\mathcal{A}}
\sum_{b\in\mathcal{B}}\widetilde{P}_{\mathbf{A},B}(\mathbf{a},b)
\log\!\left(\frac{\widetilde{P}_{\mathbf{A},B}(\mathbf{a},b)}
{\widetilde{P}_{\mathbf{A}}(\mathbf{a})\,\widetilde{P}_{B}(b)}
\right), 
\end{equation}
where \[\widetilde{P}_{\mathbf{A},B}(\mathbf{a},b) =
\frac{1}{|\mathcal{M}|} \sum_{d \in \mathcal{M}} \mathbb{1}\{\mathbf{a}_d=\mathbf{a}\}\,p_w(B_d=b),
\]
\[
\widetilde{P}_{\mathbf{A}}(\mathbf{a})
=
\frac{1}{|\mathcal{M}|}
\sum_{d\in\mathcal{M}}
\mathbb{1}\{\mathbf{a}_d=\mathbf{a}\},
\qquad
\widetilde{P}_B(b)
=
\frac{1}{|\mathcal{M}|}
\sum_{d\in\mathcal{M}} p_w(B_d=b).
\]
Adequate minibatch coverage is necessary to obtain a reliable
approximation of the distribution of $\mathbf{A}$ over
$\mathcal{A}$.

We adopt a count-based approximation rather than more computationally intensive adversarial estimation methods \cite{cho2020fair} or variational estimators (e.g., MINE \cite{belghazi2018mutual}), to avoid adversarial training, reduce computational cost, and prevent the underestimation and instability that may arise from the use of lower bounds. To ensure reliable probability estimates, the batch size must be sufficiently large and must include representative samples from all demographic subgroups. Such coverage is necessary to approximate the distribution of $\mathbf{A}$ over its full domain.

\section{Experiments}
We evaluate MIFair across multiple scenarios to assess its fairness performance. The results show that MIFair consistently achieves strong fairness performance under all tested notions, including highly intersectional settings and multiclass classification. The induced accuracy trade-offs remain modest, indicating that the method effectively adjusts the original distribution without excessive distortion.
In our experiments, we implement three of the five fairness notions introduced in Section~\ref{subsec:templatesotametric}: Statistical Parity (SP), Equal Opportunity (EO), and Overall Accuracy Equality (OAE).
\subsection{Setup}
\paragraph{\textbf{Adult Dataset}}
We evaluate MIFair on the UCI Adult dataset \cite{becker1996adult}, a standard fairness benchmark. The dataset includes 14 binary, categorical, or continuous features and a binary income label indicating whether an individual earns more than \$50K. We use a two-layer fully connected neural network with a softmax output, trained using cross-entropy loss.
We consider three sensitive attributes, \textit{race}, \textit{sex}, and \textit{relationship status}, each treated as binary, with values 
$\{\textit{White}, \textit{Non-White}\}$,
$\{\textit{Male}, \textit{Female}\}$, and
$\{\textit{Not-in-Family}, \textit{In-Family}\}$, respectively.
Non-sensitive features are \textit{age, workclass, education, education-num, marital-status, occupation, capital-gain, capital-loss, hours-per-week, native-country}.

\paragraph{\textbf{CelebA Dataset}}
We further evaluate MIFair on the large-scale CelebA vision dataset \cite{liu2015deep} using a ResNet18 model. We consider both a binary (Task 1) and a multiclass (Task 2) prediction task, namely, determining whether a person is \emph{smiling} (Task 1) and whether the person has \emph{blond}, \emph{brown}, or \emph{black} hair (Task 2), while enforcing fairness with respect to the sensitive attributes \textit{Male}/\textit{Female} and \textit{Chubby}/\textit{Non-Chubby}. The dataset is highly imbalanced, for example, the $\{\textit{Female}, \textit{Non-Chubby}\}$ subgroup contains hundreds of times as many samples as $\{\textit{Female}, \textit{Chubby}\}$. Unlike the Adult dataset, these sensitive attributes are implicit in the input (image pixels) rather than explicitly provided, demonstrating MIFair’s ability to mitigate bias in unstructured data.

Additional experimental details are deferred to Appendix \ref{sec:experimental_details}.

\paragraph{\textbf{Evaluation}}
We evaluate the performance of the trained models based on their predictions using the following two accuracy metrics:
\begin{align*}
    \operatorname{ACC_{mean}} & = \frac{1}{|D|}
\sum_{d=1}^{|D|} \mathbb{1}\{\hat{y}_d = y_d\}.\\
 \operatorname{ACC_{weighted}} & = \frac{1}{|G|} \sum_{g\in G}
\frac{1}{N_g} \sum_{d:\,\mathbf{a}_d=g}\mathbb{1}\{\hat{y}_d = y_d\},
\end{align*}
where $N_g = \left|\left\{d\in\{1,\ldots,|D|\}:\mathbf{a}_d=g\right\}\right|$. $\operatorname{ACC_{weighted}}$ assigns equal importance to each demographic subgroup's accuracy, regardless of its size, providing a clearer view of balance across subgroups. In contrast, $\operatorname{ACC_{mean}}$ maintains representation biases, as more frequent subgroups in the training data have a greater impact on the overall evaluation, potentially masking such biases. Differences of results between both metrics gives an insight of possible representation biases.

Second, we use the following \textbf{MIFair fairness metrics} for fairness evaluation:
\begin{align*}
    \operatorname{\iota_{SP}} & = I(\mathbf{A};\hat{Y}) \\
    \operatorname{\iota_{EO}} & = I(\mathbf{A};\hat{Y}\mid Y=1) \\
    \operatorname{\iota_{OAE}} & = I\!\left(\mathbf{A};\mathbb{1}\{\hat{Y}=Y\}\right).
\end{align*}

For comparison, we also employ the following \textbf{baseline metrics} from the literature. Let $(\mathbf{a},\tilde{\mathbf{a}}) \in \mathcal{A}^2$, with $\mathbf{a}\neq \tilde{\mathbf{a}}$.

\begin{align*}
\operatorname{SPD}(\mathbf{a},\tilde{\mathbf{a}}) = P(\hat{Y}=1 \mid \mathbf{A}=\mathbf{a}) - P(\hat{Y}=1 \mid \mathbf{A}=\tilde{\mathbf{a}}) \\
\operatorname{EOD}(\mathbf{a},\tilde{\mathbf{a}}) = P(\hat{Y}=1 \mid Y=1,\mathbf{A}=\mathbf{a}) - P(\hat{Y}=1 \mid Y=1,\mathbf{A}=\tilde{\mathbf{a}})\\
\operatorname{OAE}(\mathbf{a},\tilde{\mathbf{a}}) = P(\hat{Y}=Y \mid \mathbf{A}=\mathbf{a}) - P(\hat{Y}=Y \mid \mathbf{A}=\tilde{\mathbf{a}}).
\end{align*}
These metrics take values in $[-1, 1]$, as they represent differences between probabilities. Their values may be either positive or negative, depending on the ordering of the subgroups. A positive value indicates that group $\mathbf{a}$ is the privileged group, whereas a negative value indicates that $\tilde{\mathbf{a}}$ is the privileged group. A value of zero corresponds to a perfectly fair predictor under the given fairness criterion. In practice, values in the range $[-0.2, 0.2]$ are often considered acceptable \cite{saleiro2019aequitasbiasfairnessaudit}.

\paragraph{\textbf{Experiments}} 
We consider an intersectional setting with $n - m = 3$ (\textit{sex}, \textit{race}, \textit{relationship}) and $|G| = 8$ for Adult, and $n-m = 2$ (\textit{sex}, \textit{chubby}) and $|G| = 4$ for CelebA. Three experiments are conducted by training a model with the MIFair metric instantiated using $SP$, $EO$, and $OAE$, respectively. For CelebA, only Experiment~1 was conducted.

\begin{align*}
\operatorname{Exp.~1:} \quad & \min_{w\in\mathbb{R}^{h}} L(D,w) + \eta\,\iota_{SP}(D,w), \\
\operatorname{Exp.~2:} \quad & \min_{w\in\mathbb{R}^{h}} L(D,w) + \eta\,\iota_{EO}(D,w), \\
\operatorname{Exp.~3:} \quad & \min_{w\in\mathbb{R}^{h}} L(D,w) + \eta\,\iota_{OAE}(D,w).
\end{align*}

For each experiment, we verify whether the corresponding $\iota_{\mathrm{fairness}}$ metric is minimized and whether the associated baseline metric ($SPD$, $EOD$, or $OAE$) is also reduced across all subgroup pairs $(\mathbf{a},\tilde{\mathbf{a}})\in\mathcal{A}^2$, as expected from the minimization of $\iota_{\mathrm{fairness}}$ (Section \ref{subsec:templatesotametric}). For each configuration, we run five independent trials with distinct random seeds, using identical data splits across methods, and report the mean performance across runs.

\paragraph{\textbf{Varying regularization strength}}
We analyze the influence of the regularization strength by varying the value of $\eta$ in \eqref{pb: final} and identify threshold values that yield satisfactory fairness results while maintaining a favorable fairness-accuracy trade-off. $\eta=0$ (no regularization) is denoted as the ``vanilla'' setting.

\subsection{Results}\label{sec:results}

\begin{figure*}[htbp]
\centering
\begin{subfigure}[b]{0.3\textwidth}
 \centering
 \includegraphics[width=\textwidth]{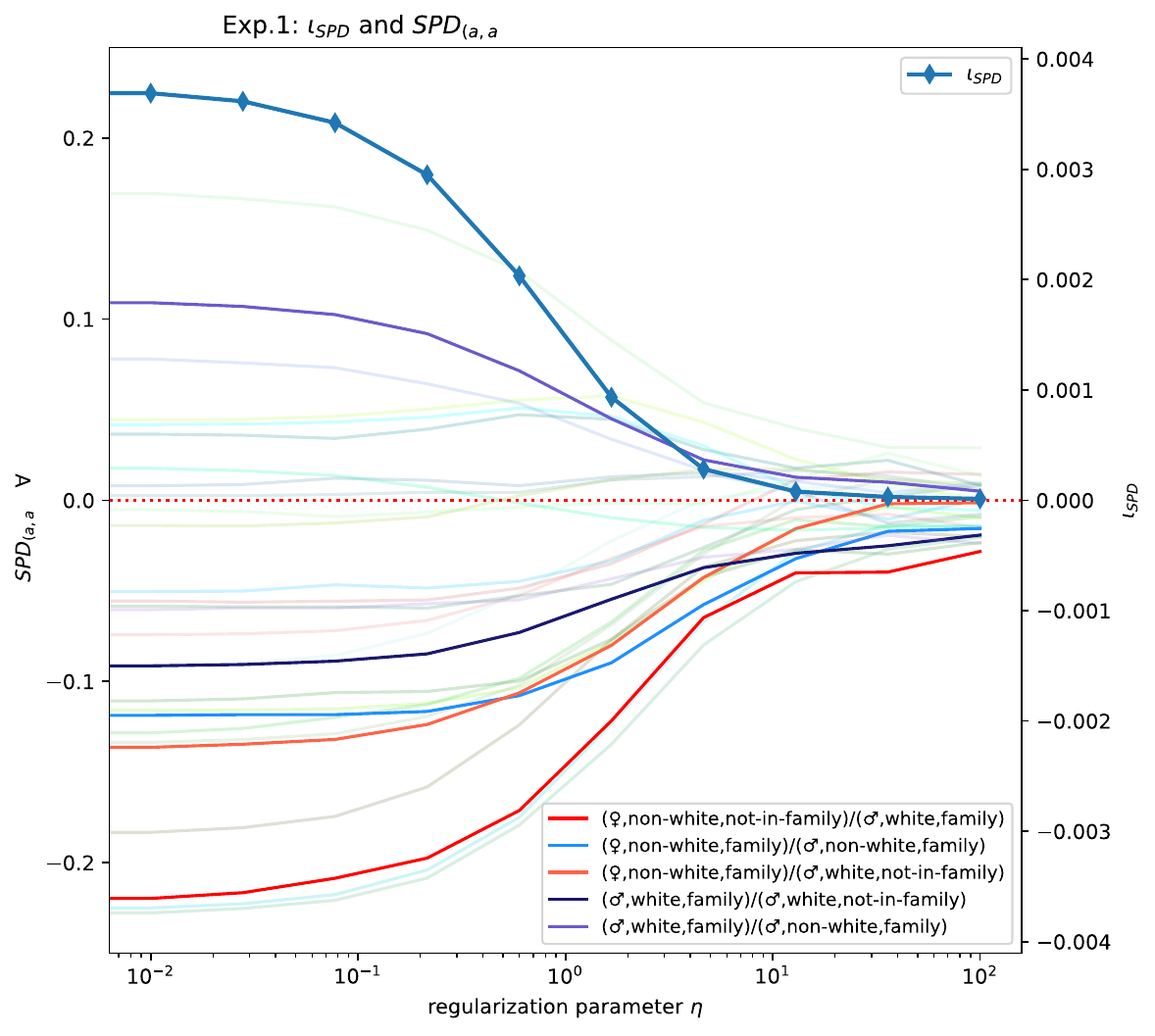}
 \subcaption{Exp.1: $SPD(\mathbf{a},\tilde{\mathbf{a}})$ \& $\mathbf{\iota}_{SP}$}
 \label{fig:exp1_spdmetrics}
\end{subfigure}
\begin{subfigure}[b]{0.3\textwidth}
 \centering
 \includegraphics[width=\textwidth]{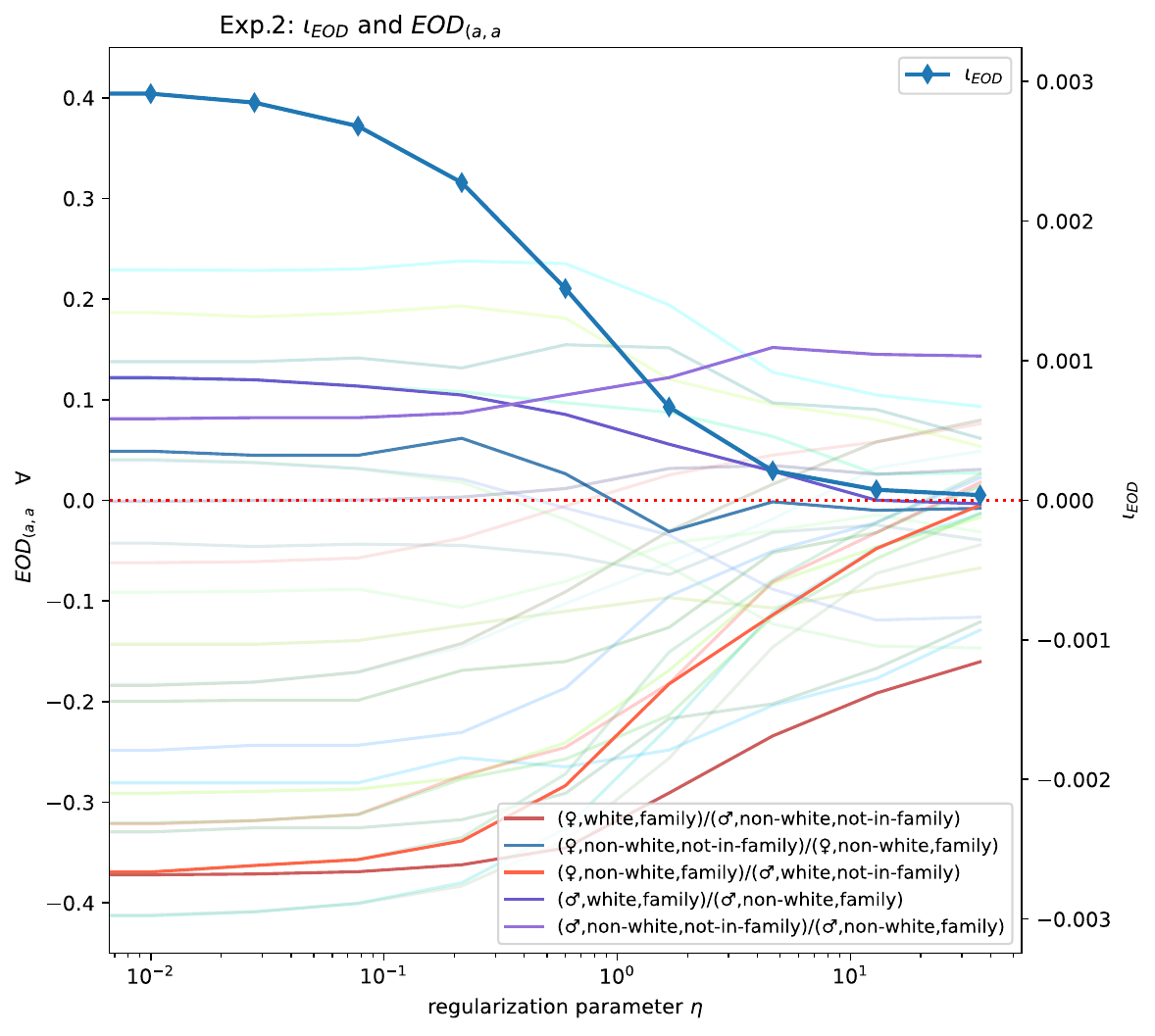}
 \subcaption{Exp.2: $EOD(\mathbf{a},\tilde{\mathbf{a}})$ \& $\mathbf{\iota}_{EO}$}
 \label{fig:exp2_eodmetrics}
\end{subfigure}
\begin{subfigure}[b]{0.3\textwidth}
 \centering
 \includegraphics[width=\textwidth]{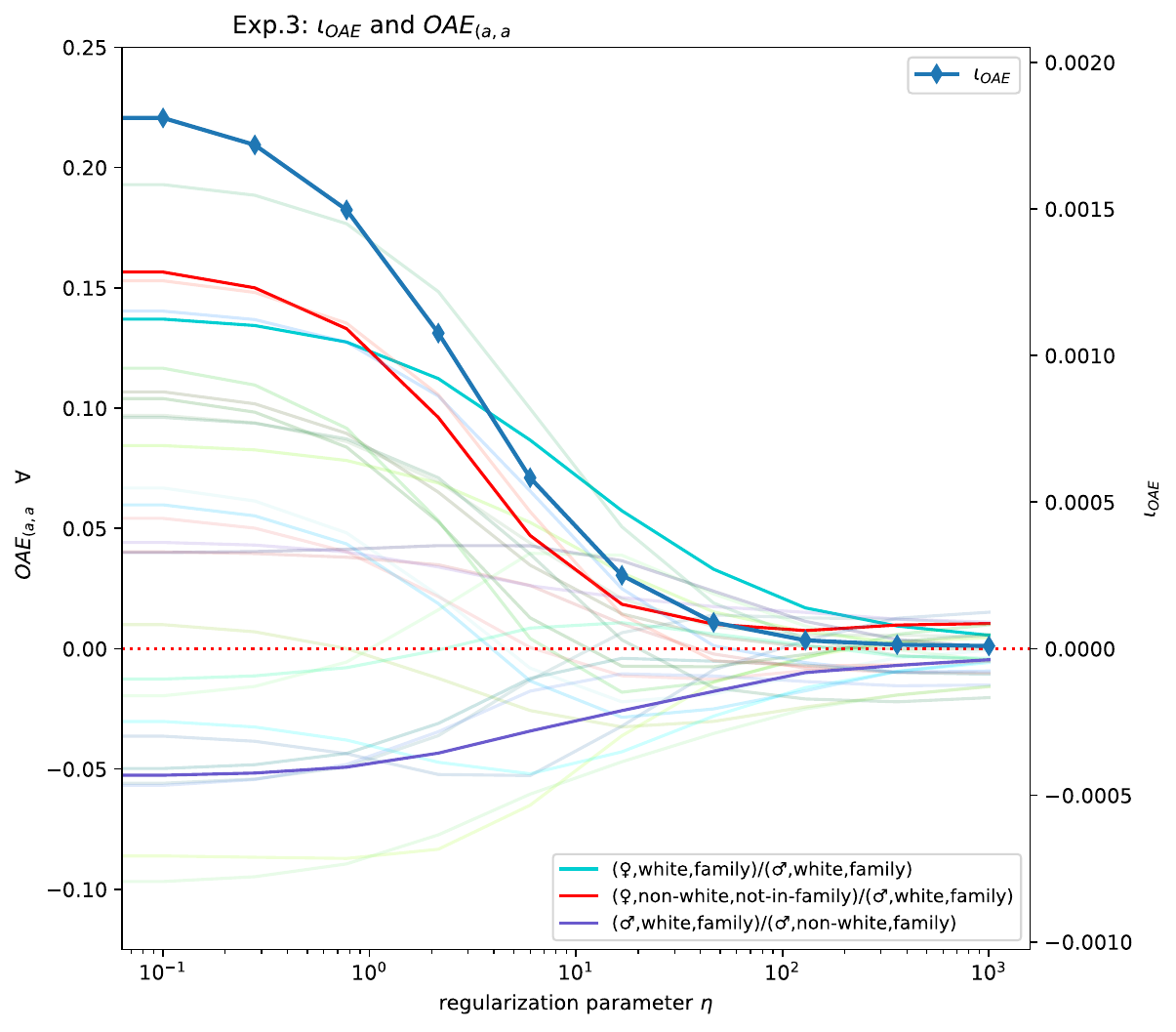}
 \subcaption{Exp.3: $OAE(\mathbf{a},\tilde{\mathbf{a}})$ \& $\mathbf{\iota}_{OAE}$}
 \label{fig:exp3_oaemetrics}
\end{subfigure}
\caption{(Adult dataset.) MIFair metric $\mathbf{\iota}_\text{fairness}$ (right y-axis) and the corresponding fairness metric over all subgroups (left y-axis) as $\eta$ varies, for three fairness definitions: $SP$ (Fig.~\ref{fig:exp1_spdmetrics}), $EO$ (Fig.~\ref{fig:exp2_eodmetrics}) and $OAE$ (Fig.~\ref{fig:exp3_oaemetrics}). The highlighted curves are discussed further in Section \ref{sec:results}.}
\label{fig:mi&sotametric}
\end{figure*}

\begin{figure}[htbp]
    \centering
     \includegraphics[width=0.35\textwidth]{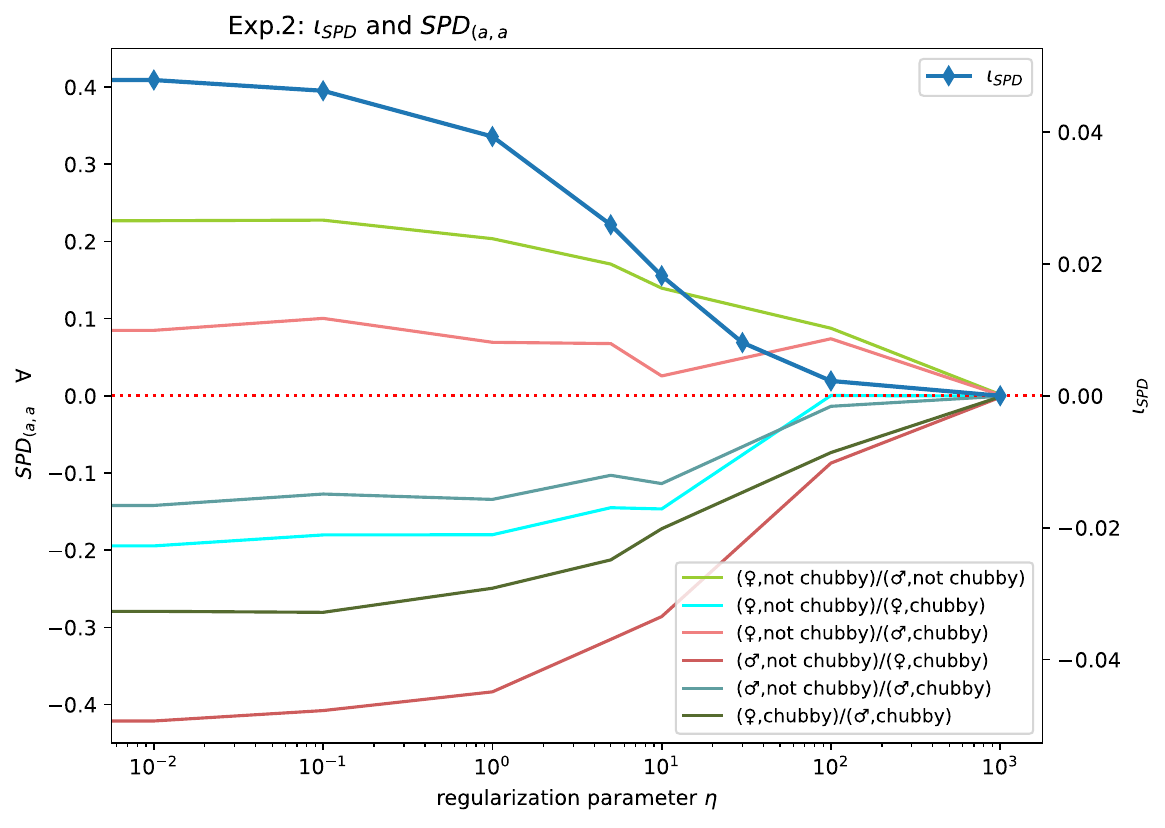}
    \caption{(CelebA dataset.) $\mathbf{\iota}_\text{SP}$ (right y-axis) and the corresponding fairness metrics from the literature over all subgroups (left y-axis) for the binary prediction task, as $\eta$ varies (x-axis) in Experiment 1 on Task 1.}
\label{fig:celeba_metrics}
\end{figure}

\paragraph{\textbf{Increased regularization strength minimizes the corresponding metric}}
As a sanity check, we verify that adding the regularization term effectively reduces the corresponding MIFair metric across fairness notions. Fig.~\ref{fig:mi&sotametric} shows that $\iota_{\text{SP}}$, $\iota_{\text{EO}}$, or $\iota_{\text{OAE}}$ decrease sharply as $\eta$ increases in Exp. 1, 2 or 3 respectively, confirming the expected effect of stronger regularization. We also observe that minimizing the MIFair term consistently reduces the corresponding classical metrics (SPD, EOD, and OAE).
For all three notions, the empirical metrics remain bounded by a decreasing function of $\eta$, demonstrating progressive bias mitigation. Using a threshold $s = 0.2$ across subgroups, we find that achieving $|SPD| \le s$ requires $\eta \ge 10^{-0.8}$, and $|EOD| \le s$ requires $\eta \ge 10^{1.1}$. The vanilla model already satisfies $|OAE| \le s$.
Fig.~\ref{fig:celeba_metrics} shows the same behavior on the CelebA dataset, confirming that MIFair performs consistently across the evaluated data distributions.

\paragraph{\textbf{MIFair handles intersectionality}}
In Fig.~\ref{fig:mi&sotametric}, the baseline metrics (SPD, EOD, OAE) are evaluated across all subgroup pairs $(\mathbf{a},\tilde{\mathbf{a}})\in \mathcal{A}^2, \mathbf{a} \ne \tilde{\mathbf{a}}$. Most values converge toward zero, which corresponds to complete fairness, as regularization increases, with a few exceptions discussed below.
Fig.~\ref{fig:exp1_spdmetrics} highlights five representative curves and shows that MIFair effectively mitigates intersectional biases across diverse subgroup differences. The two curves in shades of red-orange correspond to subgroup pairs that do not share any sensitive-attribute values, that is, to the most distinct and therefore the most biased intersectional groups. These pairs exhibit the highest disparities at $\eta = 0$. As regularization increases, their gaps steadily shrink and eventually converge toward zero, illustrating MIFair’s ability to reduce intersectional bias.
All subgroup pairs ultimately lie within the commonly accepted fairness band $[-0.2, 0.2]$. Similar behavior is observed for the red-orange curves in Figs.~\ref{fig:exp2_eodmetrics} and \ref{fig:exp3_oaemetrics}, further confirming MIFair’s robustness in intersectional settings.

\begin{figure}[htbp]
    \centering
     \includegraphics[width=0.35\textwidth]{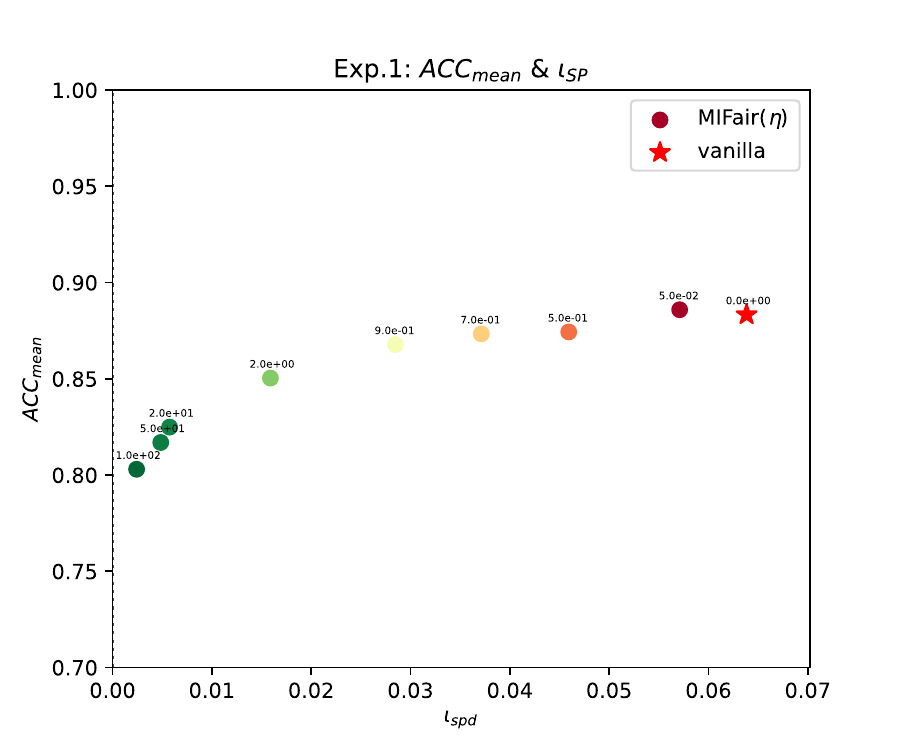}
    \caption{(CelebA dataset.) Fairness-accuracy trade-off for the multiclass classification task (Exp.1 on Task 2).}
\label{fig:celeba_tradeoff}
\end{figure}

\paragraph{\textbf{MIFair works in multiclass classification}} Fig.~\ref{fig:celeba_tradeoff} shows the results of MIFair-regularized training for a multiclass classification task under the $SP$ criterion. As observed for binary tasks, accuracy losses remain limited, with a maximum loss of $9.1\%$ under the strongest regularization setting, $\eta = 10^{2}$, whereas the fairness gains are substantial, with a $96\%$ decrease in the $\iota_{SP}$ value. Moreover, the loss in accuracy can be reduced further while maintaining a substantial level of fairness through an appropriate choice of $\eta$, thereby improving the fairness-accuracy trade-off. These results show that MIFair can effectively mitigate bias in multiclass classification tasks under the selected fairness definition.

\paragraph{\textbf{MIFair maintains performance on biases arising from binary subgroup comparisons}}
The blue and purple curves in Fig.~\ref{fig:mi&sotametric} correspond to subgroup pairs differing in only one sensitive attribute, thereby effectively reducing the comparison to a binary setting ($|G|=2$, with a single varying sensitive attribute). These curves also converge toward zero as $\eta$ increases, showing that MIFair mitigates biases in binary scenarios just as effectively as in intersectional ones.

\paragraph{\textbf{MIFair has limitations with scarce data}}
Results in Fig.~\ref{fig:exp2_eodmetrics} are slightly weaker than those in Figs.~\ref{fig:exp1_spdmetrics} and \ref{fig:exp3_oaemetrics}. As a separation-based notion, \textit{Equal Opportunity} relies only on positive-label samples, reducing the effective data available. The Adult dataset is strongly imbalanced, with negative labels outnumbering positives by roughly 3:1, which limits the accuracy of both the distribution estimates and the regularization term.
This results in suboptimal behavior for underrepresented subgroups, particularly those with $A_{\text{race}} = \text{Non-White}$, where probability estimates become less reliable, restricting MIFair’s ability to fully mitigate bias. Nevertheless, the fairness condition $|EOD| \le s$ is still satisfied, indicating that acceptable fairness performance is achieved even for these disadvantaged groups.

\begin{figure*}[h]
\centering
\begin{subfigure}[b]{0.3\textwidth}
 \centering
 \includegraphics[width=\textwidth]{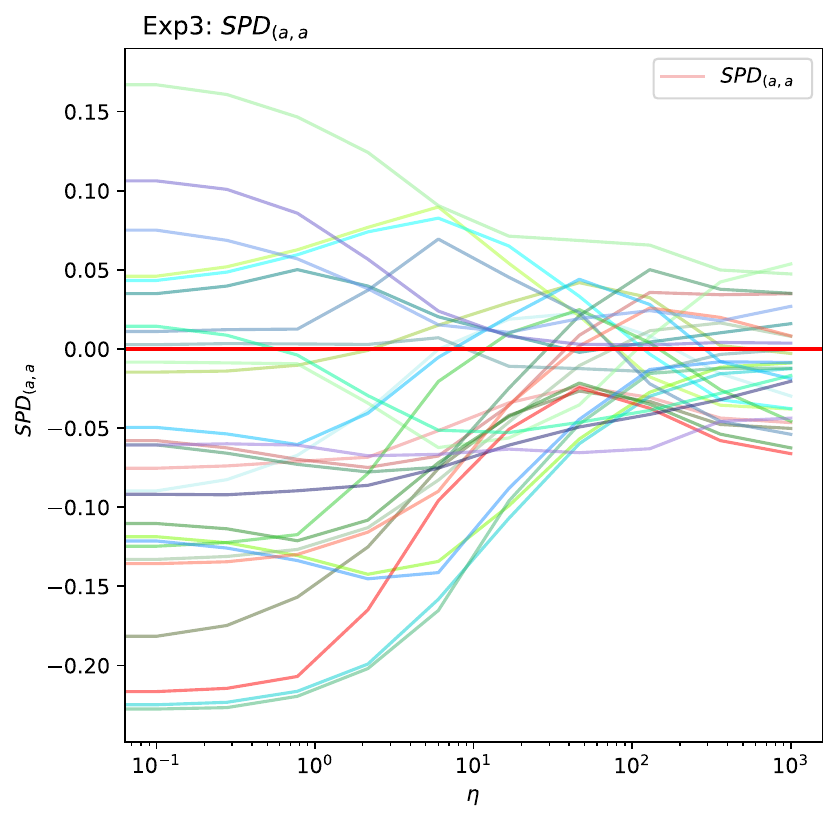}
 \caption{$SPD$ across subgroup pairs when $\mathbf{\iota}_{OAE}$ is minimized}
 \label{fig:mi_oae_sg_spd}
\end{subfigure}
\hfill
\begin{subfigure}[b]{0.3\textwidth}
 \centering
 \includegraphics[width=\textwidth]{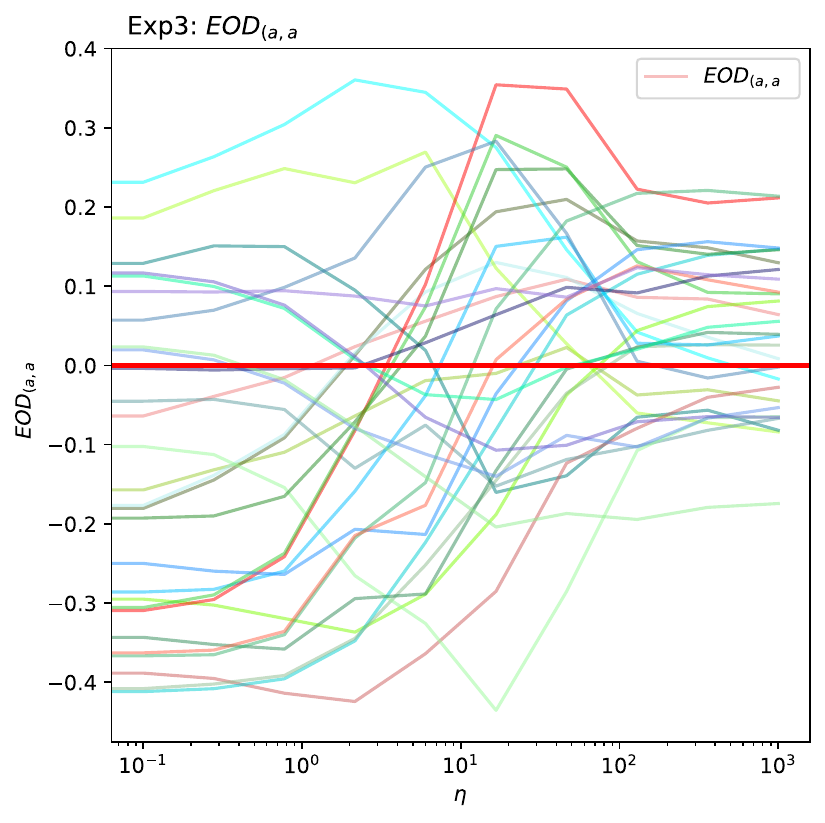}
 \caption{$EOD$ across subgroup pairs when $\mathbf{\iota}_{OAE}$ is minimized}
 \label{fig:mi_oae_sg_eod}
\end{subfigure}
\hfill
\begin{subfigure}[b]{0.3\textwidth}
 \centering
 \includegraphics[width=\textwidth]{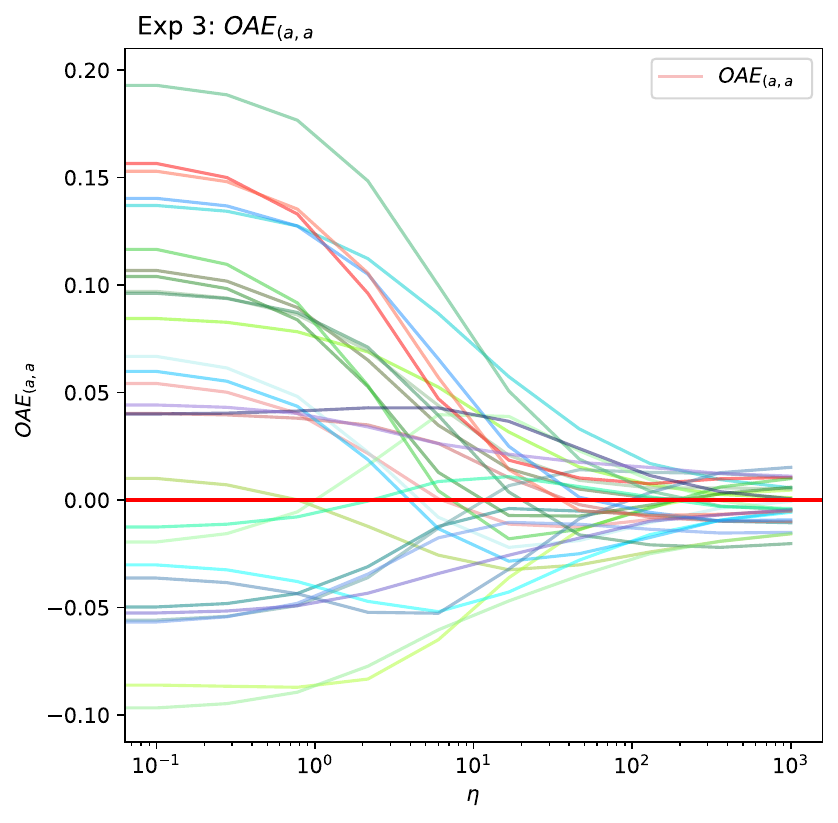}
 \caption{$OAE$ across subgroup pairs when $\mathbf{\iota}_{OAE}$ is minimized}
 \label{fig:mi_oae_sg_oae}
\end{subfigure}
\caption{(Adult dataset.) \textbf{Incompatibility between fairness notions}. 
In Exp.~3, MIFair is configured for \textit{Overall Accuracy Equality}, successfully 
minimizing $\iota_{\text{OAE}}$ and $OAE$ (Fig.~\ref{fig:mi_oae_sg_oae}). 
However, this improvement comes at the cost of degraded performance under other fairness 
notions, such as \textit{Statistical Parity} (Fig.~\ref{fig:mi_oae_sg_spd}) 
and \textit{Equal Opportunity} (Fig.~\ref{fig:mi_oae_sg_eod}), illustrating 
the inherent incompatibility between fairness criteria. Each curve corresponds to a pair of subgroups.}
\label{fig:inc_metric}
\end{figure*}

\begin{figure*}
\centering
\begin{subfigure}[b]{0.32\textwidth}
 \centering
 \includegraphics[width=\textwidth]{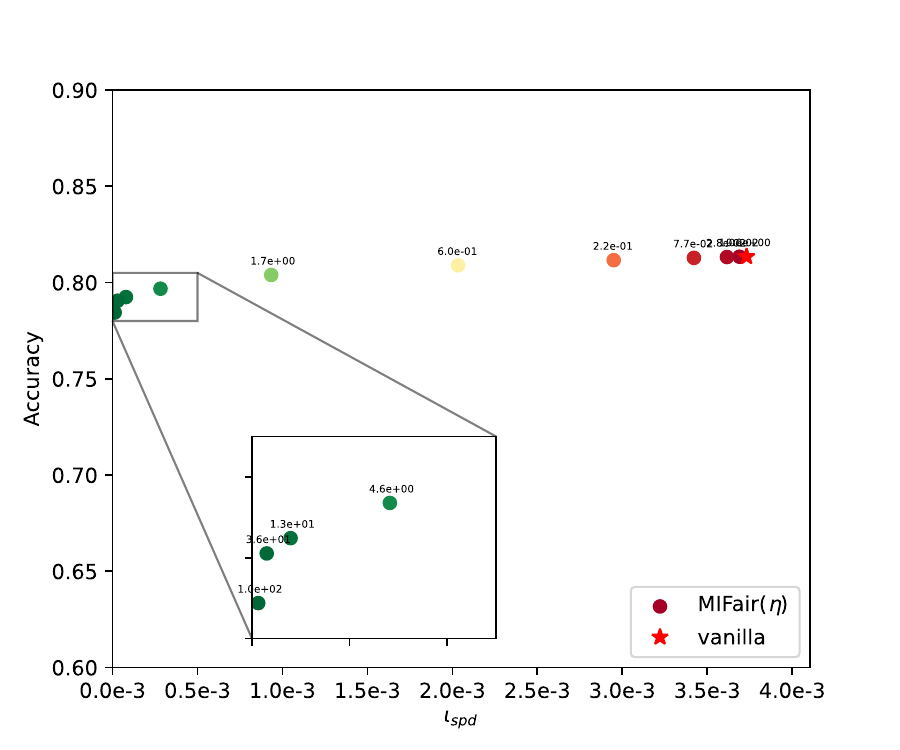}
 \caption{Exp.1: $\operatorname{ACC_{mean}}$ \& $\mathbf{\iota}_{SP}$}
 \label{fig:acc_spd}
\end{subfigure}
\hfill
\begin{subfigure}[b]{0.32\textwidth}
 \centering
 \includegraphics[width=\textwidth]{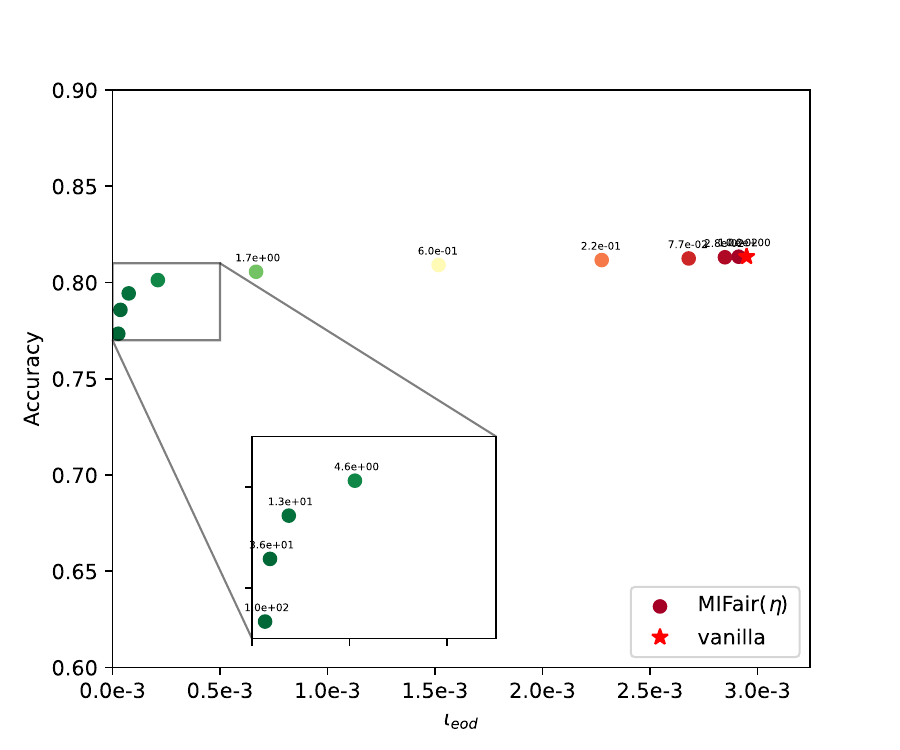}
 \caption{Exp.2: $\operatorname{ACC_{mean}}$ \& $\mathbf{\iota}_{EO}$}
 \label{fig:acc_eod}
\end{subfigure}
\hfill
\begin{subfigure}[b]{0.32\textwidth}
 \centering
 \includegraphics[width=\textwidth]{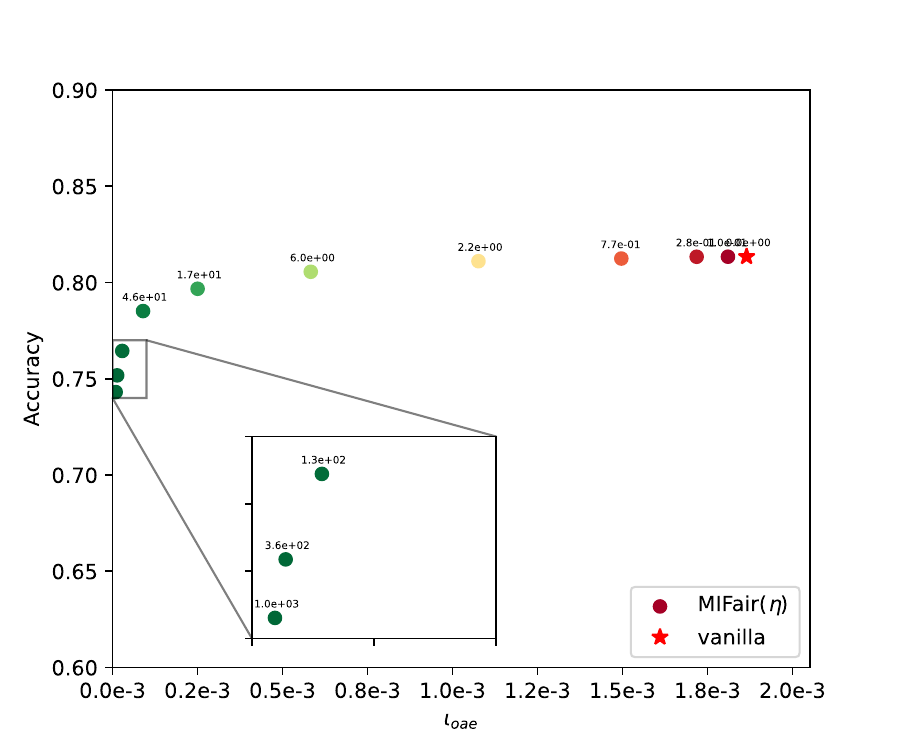}
 \caption{Exp.3:  $\operatorname{ACC_{mean}}$ \& $\mathbf{\iota}_{OAE}$}
 \label{fig:acc_oae}
\end{subfigure}
\caption{(Adult dataset.) \textbf{Fairness-accuracy trade-off}: Stronger MIFair regularization reduces the corresponding MIFair metric (x-axis) and shifts the predictions away from the original empirical distribution of the test data, as reflected by a lower value of $\operatorname{ACC_{mean}}$ (y-axis).}
\label{fig:acc_fair}
\end{figure*}

\paragraph{\textbf{Fairness notions incompatibility}}
A key advantage of MIFair is its ability to unify diverse fairness definitions within a single framework. To illustrate the consequences of selecting an unsuitable fairness notion, Fig.~\ref{fig:inc_metric} shows the evolution of three metrics in Experiment~3 on Adult, where only $\iota_{\text{OAE}}$ is minimized, targeting \textit{Overall Accuracy Equality}.
While Fig.~\ref{fig:mi_oae_sg_oae} shows that $OAE$ effectively converges to zero across all subgroup pairs as $\eta$ increases, Figs.~\ref{fig:mi_oae_sg_spd} and \ref{fig:mi_oae_sg_eod} reveal that $SPD$ and $EOD$ are not minimized under this regularization, unlike in Experiments~1 and 2 (Fig.~\ref{fig:mi&sotametric}). $SPD$ shows only minor improvement, far below the gains obtained in Experiment~1, and $EOD$ is substantially degraded, offering no fairness benefit and in some cases worsening disparities.
These findings highlight the inherent incompatibility between fairness notions and underscore the importance of selecting the correct metric for the fairness objective. MIFair mitigates this risk by providing a unified, accessible framework that guides practitioners toward appropriate bias mitigation choices.

\paragraph{\textbf{A unified benchmark for comparing disparate fairness notions}} MIFair enables bias comparison across different fairness notions using a single measure, even in intersectional contexts. In Figure \ref{fig:mi&sotametric}, $\mathbf{\iota}_{fairness}$ is plotted for each experiment. Depending on the fairness definition, the MIFair metric starts at different initial values when $\eta =0$: $\mathbf{\iota}_{SP}=0.0037$ (Exp.1),  $\mathbf{\iota}_{EO}=0.0029$ (Exp. 2), and $\mathbf{\iota}_{OAE}=0.0019$ (Exp. 3). Thanks to MIFair's standardization given by the mutual information-based approach, this single value allows for direct comparison of bias across demographic subgroups, regardless of their number. For instance, the vanilla model exhibits less bias regarding \textit{Overall Accuracy Equality} than \textit{Statistical Parity}, a finding corroborated by the classical metrics measured for subgroup pairs.

\paragraph{\textbf{Fairness-accuracy trade-off}}
We analyze how accuracy evolves under increasing regularization. Accuracy reflects how much information the model extracts from the data; if it collapses, training becomes meaningless. However, when fairness is prioritized, the relevance of accuracy, especially  $\operatorname{ACC_{mean}}$, depends on the fairness notion being enforced. Achieving fairness often requires modifying the original data distribution; therefore, some decrease in accuracy is to be expected and may, in certain cases, even be desirable.
\begin{figure}[h]
    \centering
     \includegraphics[width=0.33\textwidth]{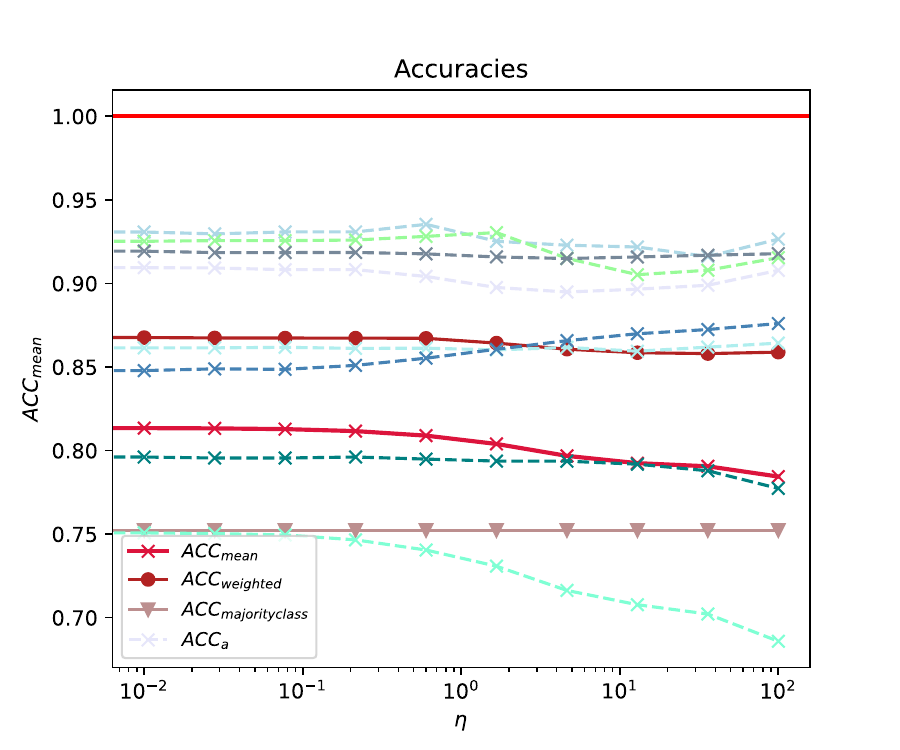}
    \caption{(Adult dataset.) Exp. 1: Accuracy decreases slightly as $\eta$ increases.}
\label{fig:accuracies}
\end{figure}

In Section~\ref{sec:intro}, we noted that representation bias causes underrepresented subgroups to contribute less to the objective, making $\operatorname{ACC_{mean}}$ a poor indicator of subgroup-level performance. In contrast, $\operatorname{ACC_{weighted}}$ better captures accuracy across subgroups. When representation biases are reduced, $\operatorname{ACC_{weighted}}$ may increase, or decline more slowly, even as  $\operatorname{ACC_{mean}}$ decreases.
Fig.~\ref{fig:accuracies} illustrates this behavior in Experiment~1, where regularization targets $\iota_{\text{SP}}$. As regularization increases, $\operatorname{ACC_{weighted}}$ drops less sharply than  $\operatorname{ACC_{mean}}$, indicating a partial correction of representation bias. For reference, we also report accuracy when trivially predicting the majority class as a baseline sanity check.

The fairness-accuracy trade-off and its evolution with $\eta$ reflect how closely the model follows the original, potentially biased, data distribution and how much accuracy is sacrificed to improve fairness. Figs.~\ref{fig:acc_fair} and \ref{fig:celeba_tradeoff} show the corresponding gains in fairness and losses in accuracy. Across all experiments, $\iota_{\text{fairness}}$ decreases substantially, whereas the drop in accuracy at the largest values of $\eta$ remains limited relative to the vanilla model.
Although we intentionally consider large values of $\eta$ to strongly reduce the fairness metrics, such extreme regularization is generally unnecessary in practice; moderate values of $\eta$ typically provide a more favorable fairness-accuracy trade-off.

\paragraph{\textbf{Baselines}} Comparing our method with prior work is nontrivial because most baselines neither support nor evaluate intersectional settings or multiclass outputs. Existing code and protocols typically assume one binary sensitive attribute and binary classification, so a like-for-like comparison would either narrow our study and weaken our main contribution or require substantial third-party extensions that would confound the comparison. Furthermore, MIFair adopts an information-theoretic evaluation framework that quantifies the dependence between the sensitive-attribute vector $\mathbf{A}$ and a notion-specific variable $B$ using mutual information, rather than pairwise subgroup disparity metrics such as SPD, EOD, and OAE. Although both approaches coincide at zero bias, they capture different forms of residual dependence; evaluating competing methods under a single framework would therefore systematically favor that framework.

Accordingly, in Table~\ref{tab:mifair-vs-baseline}, we compare MIFair with the KDE-based framework proposed by \cite{cho2020fairkde} on two fairness criteria. This framework aims to reduce subgroup deviations from the overall positive prediction rate with respect to demographic parity:
\begin{equation}
DDP := \sum_{\mathbf{a}\in\mathcal{A}} \left|\Pr(\hat{Y}=1 \mid \mathbf{A}=\mathbf{a}) - \Pr(\hat{Y}=1) \right|.
\end{equation}

We conducted Exp. 1 on Adult with $n-m = 2$ (\textit{sex}, \textit{race}) using the authors' code from the supplementary material for the KDE-based framework. For the comparison with~\cite{cho2020fairkde}, we use the preprocessing and subgroup definition provided in the authors' code, which differs from the binary race encoding used in our main Adult experiments and yields $|G|=20$.

We report both criteria, $\iota_{SP}$ and DDP, across all experiments. As expected, MIFair achieves a lower $\iota_{SP}$, whereas the KDE-based baseline yields a smaller DDP. However, MIFair exhibits substantially better accuracy while still tending toward a virtually unbiased model.
Beyond performance, this comparison highlights that mutual information and pairwise distributional metrics capture bias differently. More broadly, especially in intersectional settings, it shows that aggregating subgroup-level measures into a single fairness score can materially affect conclusions about residual bias \cite{mon2026beyond}.

\begin{table}[t]
  \centering

  \begin{tabular}{|c||c|c|c||c|c|c|}
  \hline {$\eta/\eta_{max}$} & \multicolumn{3}{c}{MIFair} & \multicolumn{3}{c}{Baseline \cite{cho2020fairkde}} \\
\hline
\hline
       & {Acc.} & {DDP} & {MI} & {Acc.} & {DDP} & {MI} \\
\hline
    $0$        & \textbf{0.846} 
                             & 1.10 
                             & $2.83\cdot10^{-2}$ 
                             & \textbf{0.846} 
                             & 1.10 
                             & $2.83\cdot10^{-2}$ 
    \\
    \hline
    $0.5$  & \textbf{0.834}
                             & 0.610
                             & $0.029$ 
                             & 0.760 
                             & 0.021
                             & $0.32$
    \\
    \hline
    $0.75$   & 0.827
                             & 0.541
                             & $0.012$
                             & 0.676
                             & 0.086
                             & $0.035$
    \\
    \hline
    $0.9$   & 0.788
                             & 0.191
                             & \textbf{$0.0045$}
                             & 0.602
                             & \textbf{0.081}
                             & $0.0054$
    \\
    \hline
  \end{tabular}
    \caption{(Adult dataset.) Comparative evaluation of MIFair and the baseline \cite{cho2020fairkde} across regularization strengths $\eta$ in Experiment 1.}
  \label{tab:mifair-vs-baseline}
\end{table}

\section{Conclusion}
We introduced MIFair, a unified mutual-information-based framework for assessing and mitigating bias across various fairness notions previously proposed in the AI/ML literature  and originally defined in binary settings, thereby extending their applicability to intersectional and multiclass scenarios. The resulting regularization-based in-processing method supports multiple sensitive-attribute configurations and prediction settings. Through extensive experiments on neural-network models for tabular data and on deep models for image classification and a detailed analysis of results, we demonstrated that MIFair provides a robust and versatile fairness mechanism. It substantially improves fairness by reducing bias while incurring only limited accuracy loss. Future work includes exploring alternative mutual-information estimation methods and extending MIFair to continuous features and outputs through continuous mutual-information formulations.

\bibliography{mybibfile}
\bibliographystyle{IEEEtran}


\newpage

\section{APPENDIX}

\subsection{Proofs of equivalence between fairness notions and MIFair formulation}\label{sec:proofs}
\begin{proof}
\textit{(Statistical Parity)} Suppose that we have $I(\mathbf{A};\hat{Y})=0$. Then, assuming $P(\mathbf{A})\neq 0$: $P(\mathbf{A} ; \hat{Y}) = P(\hat{Y}) P(\mathbf{A})$ $\Leftrightarrow$ $P(\hat{Y} |\mathbf{A})P(\mathbf{A}) = P(\hat{Y})P(\mathbf{A})$ $\Leftrightarrow$ $P(\hat{Y}|\mathbf{A}) = P(\hat{Y})$
and $\forall (a_1, a_2) \in A^2$, $P(\hat{Y} \mid \mathbf{A} = a_1) = P(\hat{Y} \mid \mathbf{A} = a_2) = P(\hat{Y})$. In particular, when demographic groups are defined in binary terms, we have $P(\hat{Y} \mid \mathbf{A} = 1) = P(\hat{Y} \mid \mathbf{A} = 0)$.
\end{proof}

\begin{proof}
\textit{(Equal Opportunity)} Suppose that we have $I_{Y=1}(\mathbf{A};\hat{Y})=0$. Then, assuming $P_{Y=1}(\mathbf{A})\neq 0$, we have 
\begin{equation*}
    P_{Y=1}(\mathbf{A} ; \hat{Y}) = P_{Y=1}(\mathbf{A}) P_{Y=1}(\hat{Y}) \Leftrightarrow  P_{Y=1}(\hat{Y} \mid \mathbf{A})=P_{Y=1}(\hat{Y})
\end{equation*}
and $\forall (a_1, a_2) \in A^2,$
\begin{align*}
  P_{Y=1}(\hat{Y}\mid\mathbf{A}=a_{1}) & = P_{Y=1}(\hat{Y}\mid\mathbf{A}=a_{2}) \\ \Leftrightarrow \frac{P_{Y=1}(\hat{Y};\mathbf{A}=a_{1})}{P_{Y=1}(\mathbf{A}=a_{1})} & =\frac{P_{Y=1}(\hat{Y};\mathbf{A}=a_{2})}{P_{Y=1}(\mathbf{A}=a_{2})} \\
   \Leftrightarrow \frac{P(\hat{Y};Y=1;\mathbf{A}=a_{1})}{P(Y=1;\mathbf{A}=a_{1})} & =\frac{P(\hat{Y};Y=1;\mathbf{A}=a_{2})}{P(Y=1;\mathbf{A}=a_{2})} \\ \Leftrightarrow P(\hat{Y}|Y=1;\mathbf{A}=a_{1})&  =P(\hat{Y}|Y=1;\mathbf{A}=a_{2}) \\ \Rightarrow P(\hat{Y}=1|Y=1;\mathbf{A}=a_{1}) & =P(\hat{Y}=1|Y=1;\mathbf{A}=a_{2}).
\end{align*}
In particular, when demographic groups are defined in binary terms: $P(\hat{Y} = 1|Y = 1;\mathbf{A} = 1) = P(\hat{Y} = 1|Y = 1;\mathbf{A} = 0)$.
\end{proof}

\begin{proof}
\textit{(Predictive Equality)} Following the same reasoning as for Equal Opportunity, suppose that we have $I_{Y=1}(\mathbf{A};\hat{Y})=0$. Then, $\forall (a_1, a_2) \in A^2$, $P(\hat{Y} = 1 | Y = 0 ; \mathbf{A} = a_1) = P(\hat{Y} = 1 | Y = 0 ; \mathbf{A} = a_2)$.
In particular, when demographic groups are defined in binary terms: $P(\hat{Y} = 1 | Y = 0; \mathbf{A} = 1) = P(\hat{Y} = 1 | Y = 0; \mathbf{A} = 0)$.
\end{proof}

\begin{proof}
\textit{(Equalized odds)} The proof proceeds similarly to the case of Equal Opportunity and Predictive Equality.
\end{proof}

\begin{proof} 
\textit{(Overall Accuracy Equality)} Suppose that we have $I(\mathbf{A}; (\hat{Y} = Y)) = 0$. Then, assuming $P(\mathbf{A})\neq0)$, we have: $P(\mathbf{A};(\hat{Y} = Y)) = P(\mathbf{A}) P((\hat{Y} = Y)$ $\Leftrightarrow$ $P((\hat{Y} = Y) \mid \mathbf{A})P(\mathbf{A}) = P((\hat{Y}=Y))P(\mathbf{A})$ $\Leftrightarrow$ $P((\hat{Y} = Y) \mid \mathbf{A}) = P((\hat{Y} = Y)$.
In fine, $I(\mathbf{A};(\hat{Y}=Y)=0 \Leftrightarrow P(\hat{Y} = Y \mid A) = P(\hat{Y} = Y)$ and $\forall (a_1, a_2) \in A^2$, $P(\hat{Y} = Y \mid \mathbf{A} = a_1) = P(\hat{Y} = Y \mid \mathbf{A} = a_2) = P(\hat{Y} = Y)$.

In particular, when demographic groups are defined in binary terms: $P(\hat{Y} = Y \mid \mathbf{A} = 1) = P(\hat{Y} = Y \mid \mathbf{A} = 0).$
\end{proof} 

\subsection{Additional experimental details}\label{sec:experimental_details}
\subsubsection{Adult}
We use a two-layer fully connected neural network with one hidden layer of 16 neurons and a softmax output, trained using cross-entropy loss. Training runs for up to 500 epochs, using the full training set as a single batch to ensure adequate subgroup representation. We employ SGD with momentum $\mu = 0.8$ and an L2 weight decay of $0.1$.
For the regularization hyperparameter $\eta$, we sweep 10 exponentially spaced values: $10^{-2}$--$10^{2}$ for SP and EO experiments, and $10^{-1}$--$10^{3}$ for OAE, the latter requiring larger values due to the smaller magnitude of $\iota_{\text{OAE}}$. We also include a baseline ``vanilla model’’ with $\eta = 0$. The learning rate is decreased from $10^{-1}$ to $10^{-2}$ as $\eta$ increases.
After filtering records with missing values, the dataset contains 45{,}222 instances, split into 75\% training (33{,}916 samples) and 25\% test (11{,}306 samples). The label distribution is imbalanced (75\% negative, 25\% positive), and demographic subgroup proportions range from 48\% in overrepresented groups to 2\% in underrepresented ones.
\newline

\subsubsection{CelebA}
CelebA contains $202,600$ face images with metadata providing numerous attributes, including sensitive ones. We consider two binary sensitive attributes—\textit{Male} and \textit{Chubby}—and predict the \textit{Smiling} attribute in Task 1 and \textit{Blond Hair}, \textit{Brown Hair} and \textit{Black Hair} attributes that we encode in one multi categorical attribute \textit{Hair color} in Task 2. We fine-tune a ResNet18 model from PyTorch for 15 epochs, using $8,000$ training images per epoch. The dataset is highly imbalanced across demographic groups; for example, the ${Female, Non-Chubby}$ subgroup has hundreds of times more samples than the ${Female, Chubby}$ subgroup.


\end{document}